\documentclass[10pt,conference]{IEEEtran}
\usepackage[T1]{fontenc}
\usepackage{amsmath,amssymb,amsfonts}
\usepackage{graphicx}
\usepackage{cite}
\usepackage{booktabs}
\usepackage{multirow}
\usepackage{url}
\usepackage[hidelinks,breaklinks=true]{hyperref}
\begin{document}

\newcommand{\acceptedvenue}{Accepted for publication in the Proceedings of the 2027 IEEE Symposium Series on Computational Intelligence (IEEE SSCI 2027)}

\title{\vspace*{-1.5cm}%
{\normalfont\normalsize\underline{\acceptedvenue}}\\[2.5ex]
Click and Count: Multimodal Prototype Adaptation for Class-Agnostic Object Counting}

\author{\IEEEauthorblockN{\textbf{Amani Aifi}\textsuperscript{1}, \textbf{Wassim Bouachir}\textsuperscript{1}, \textbf{Ahmed Zgaren}\textsuperscript{1} and \textbf{Nizar Bouguila}\textsuperscript{2}}
\IEEEauthorblockA{\textsuperscript{1}Universit\'e T\'ELUQ, Montr\'eal, Qu\'ebec, Canada \\
\textsuperscript{2}Concordia University, Montr\'eal, Qu\'ebec, Canada \\
aifi.amani@univ.teluq.ca, \{Wassim.Bouachir, Ahmed.Zgaren\}@teluq.ca, nizar.bouguila@concordia.ca}
}

\maketitle

\begin{abstract}
We propose a multimodal, class-agnostic, densitybased object counting framework that specifies the object of
interest through visual exemplars, a text description, or both,
within a single iterative refinement pipeline. Existing counters
typically fall into two groups: methods that fuse two modalities,
a hand-drawn bounding box and a text description, to specify
the target object, and single-modality methods that rely on either
a box exemplar or text alone. A few recent methods replace
the box with a point or click. We expand a single click into
a precise object mask via a frozen visual-prompt segmentation
model, feeding an iterative prototype-refinement process that also
fuses a text description at every refinement step. We make two
contributions: first, a click-derived exemplar extraction module
that removes the bounding-box requirement of prior exemplarbased counters; second, a per-step text-fusion mechanism that
combines visual and textual prompts throughout refinement.
Combined with text guidance, our model achieves state-of-theart counting accuracy, outperforming prior click-based counting
methods across all metrics and surpassing the box-exemplar base
architecture in mean absolute error.
The code is available at
\href{https://github.com/ameniifi/Click-and-Count}{\textbf{github.com/ameniifi/Click-and-Count}}.
\end{abstract}

\begingroup
\renewcommand{\thefootnote}{}
\footnotetext{
© 2027 IEEE. Personal use of this material is permitted.  Permission from IEEE must be obtained for all other uses, in any current or future media, including reprinting/republishing this material for advertising or promotional purposes, creating new collective works, for resale or redistribution to servers or lists, or reuse of any copyrighted component of this work in other works.
}
\endgroup
\section{Introduction}
\label{sec:intro}

Object counting is a fundamental computer vision task that aims to estimate the
number of instances of a target object category in an image. It plays an
important role in applications such as crowd analysis, industrial inspection,
agriculture monitoring, and biological image analysis. Traditional counting
approaches typically learn category-specific representations from annotated
datasets \cite{Zhang2016MCNN,Li2018CSRNet}, which limits their ability to
generalize to novel object categories without additional training or
annotations.

To address this limitation, class-agnostic object counting has emerged as a
more flexible formulation \cite{Lu2018GeneralizedFewShotCounting,Ranjan2021LearningToCount},
where the target object category is specified at
inference time through additional guidance. Instead of relying on a fixed set
of categories, the model must identify and count arbitrary objects based on
user-provided information, such as visual examples, semantic descriptions, or
spatial prompts. However, providing effective guidance remains a key challenge,
as the quality of the object representation directly influences the counting
performance.

\begin{figure}[t]
    \centering
    \includegraphics[width=\linewidth]{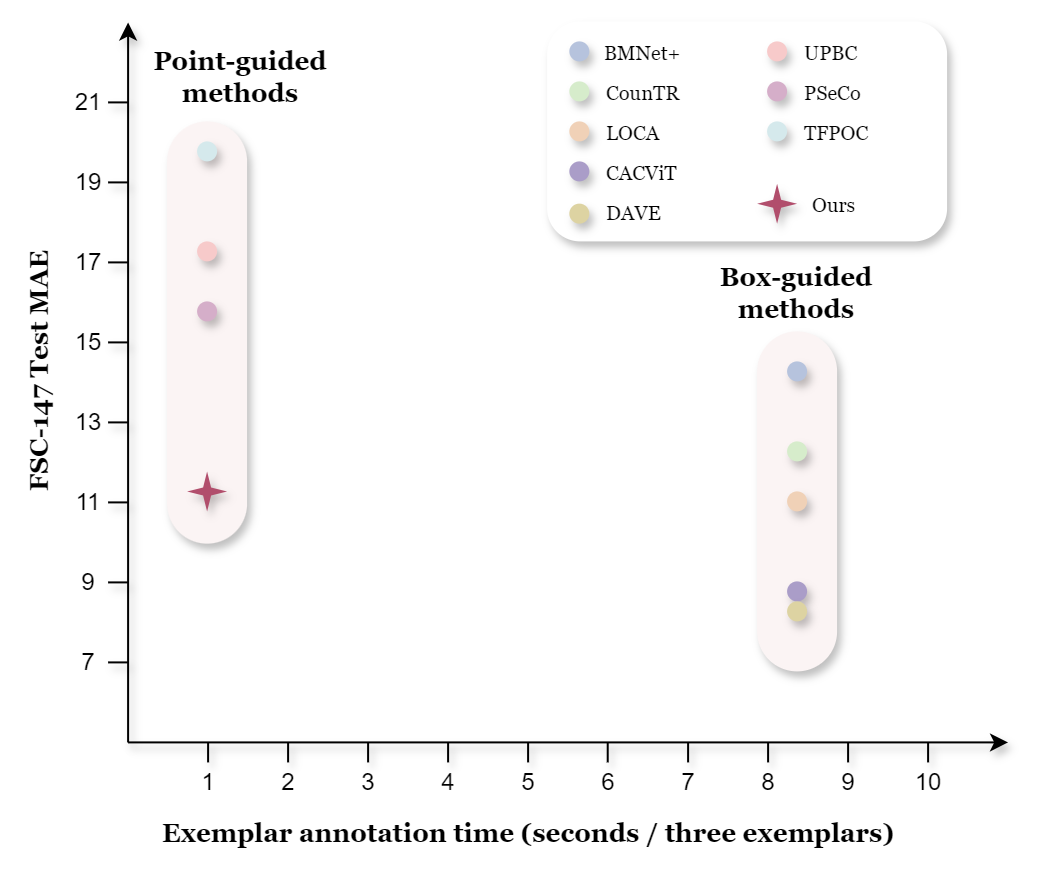}
    \caption{Annotation time is estimated by averaging the time required to
annotate three exemplars on randomly selected samples. This comparison
considers only visual exemplar annotations (clicks and bounding boxes) and
does not include text prompts. Click-based annotations substantially reduce
annotation effort while maintaining competitive counting performance.
}
    \label{fig:annotation_efficiency}
\end{figure}

A widely used paradigm for class-agnostic counting relies on visual exemplars
\cite{Shi2022BMNet,Dukic2023LOCA},
where users provide examples of the target object to guide the model. These
exemplars provide valuable instance-level information, including object
appearance and spatial characteristics. However, obtaining such exemplars
usually requires precise object localization, often through manually defined
bounding boxes. Although bounding boxes provide rich spatial supervision,
creating multiple tight annotations is time-consuming and limits the
scalability of exemplar-based counting systems.

As illustrated in Fig.~\ref{fig:annotation_efficiency}, click annotations offer
a significantly simpler alternative by requiring only a single point per
exemplar indicating
the target object location. However, clicks provide substantially less
information than bounding boxes and do not directly describe the complete
object extent. Therefore, a key challenge is how to transform sparse point
interactions into informative object representations while preserving the
advantages of exemplar-based counting.

Another challenge concerns the use of semantic information. Text descriptions
\cite{Xu2023ZSOC,Jiang2023CLIPCount}
provide category-level knowledge without requiring manual visual annotations,
but they lack instance-specific details such as appearance, scale, and spatial
structure. Consequently, visual and semantic information provide complementary
but incomplete descriptions of the target objects. Effectively combining these
two modalities remains an important direction for improving generalizable
object counting.

In this work, we introduce C\&C (Click-and-Count), a multimodal click-guided
counting framework that reformulates exemplar-based object counting with
minimal user interaction. Instead of requiring manually specified bounding
boxes, C\&C uses sparse clicks together with a click-driven object extraction
process to recover object-aware visual representations. These representations
capture both spatial and appearance information, enabling iterative prototype
refinement from lightweight user annotations.

Our main contributions are summarized as follows:

\begin{itemize}
\item We introduce C\&C (Click-and-Count), a click-guided class-agnostic
counting framework that enables exemplar-based counting using sparse point
annotations instead of manually defined bounding boxes.
\item We propose a click-driven exemplar representation strategy that converts
minimal user interaction into object-aware visual prototypes containing spatial
and appearance information.
\item We introduce semantic prototype conditioning as an optional multimodal
extension, allowing textual information to complement visual prototypes during
iterative refinement.
\item Extensive experiments demonstrate that C\&C achieves competitive
counting performance while requiring substantially simpler user interaction,
and generalizes effectively to unseen object categories.
\end{itemize}

Code and data to reproduce our results are available at
\url{https://github.com/ameniifi/Click-and-Count}.

\section{Related Work}
\label{sec:related}

Counting methods are either detection-based, localizing individual instances
with boxes or masks \cite{Ren2015FasterRCNN,He2017MaskRCNN}, or density-based,
integrating a predicted density map to obtain the count
\cite{Zhang2016MCNN,Li2018CSRNet}; the latter is more robust to occlusion and
crowding. Class-agnostic counting extends this idea to categories specified
only at inference time. The FSC-147 benchmark
\cite{Ranjan2021LearningToCount} established the standard protocol for this
setting and motivated exemplar-based methods that match bounding-box
exemplars against image features, including GMN
\cite{Lu2018GeneralizedFewShotCounting}, FamNet
\cite{Ranjan2021LearningToCount}, BMNet \cite{Shi2022BMNet}, SPDCN
\cite{Lin2022SPDCN}, CounTR \cite{Liu2022CounTR}, and LOCA \cite{Dukic2023LOCA}. Bounding-box exemplars also guide the
few-shot variant of DAVE \cite{Pelhan2024DAVE}, which verifies candidate
detections and uses the verified detections to refine its density map.
Across these approaches, the boxes specify which objects to count. We
retain this role for visual exemplars in our density-based framework
and construct them from single clicks.

More recent work reduces annotation cost with prompts beyond bounding boxes.
Text-guided methods (ZSOC \cite{Xu2023ZSOC}, CLIP-Count
\cite{Jiang2023CLIPCount}, GroundingREC \cite{GroundingREC2024}) exploit
vision-language pretraining but lack instance-specific appearance cues.
Point-based methods trade a box for a click: PSeCo \cite{Huang2024PseCo} and
TFPOC \cite{Shi2024TFPOC} explore point-supervised and segmentation-based
counting, while UPBC \cite{Lin2024UPBC} unifies box, point, and text prompts
into a shared representation. Unlike these methods, we expand each user-provided click into a
mask-derived shape and appearance prototype and fuse text at every
iterative refinement step.

\section{Method}

\subsection{Problem Formulation}

Given an input image
\begin{equation}
I \in \mathbb{R}^{3\times H\times W},
\end{equation}
where $H$ and $W$ denote the image height and width, respectively, the goal of
exemplar-based object counting is to predict the spatial distribution of all
instances belonging to the target object category.

The model receives a set of exemplar specifications describing objects of
interest:
\begin{equation}
C=\{c_i\}_{i=1}^{N},
\qquad
c_i=(x_i,y_i),
\end{equation}
where each $c_i$ denotes the spatial coordinates of a single user click
indicating one example object instance, and $N=3$ is the number of exemplars
(one click each), following the FSC-147 protocol.

From these exemplars, the model predicts a density map:
\begin{equation}
R\in\mathbb{R}^{H\times W},
\end{equation}
where the estimated object count is obtained by:
\begin{equation}
\hat{N}=\sum_{x,y}R(x,y).
\label{eq:count}
\end{equation}

Our framework is a multimodal iterative prototype-based object counting method
designed to minimize exemplar annotation while preserving accurate object
representation. Instead of requiring manually drawn bounding boxes, the proposed
framework relies on a semi-automatic exemplar extraction mechanism that combines
a single user click per exemplar with a promptable visual segmentation model to recover the
complete object region. The predicted object mask is subsequently used to
extract complementary shape and appearance prototypes, providing rich visual
representations with minimal human intervention.

In addition to visual information, the framework incorporates semantic guidance
through a pretrained vision-language model. Semantic features are injected into
the visual prototypes before and during iterative refinement, producing
multimodal object prototypes that jointly encode appearance, shape, and
category-level semantics for robust class-agnostic object counting.

\subsection{Overview of the Proposed Framework}

\begin{figure*}[t]
    \centering
    \includegraphics[width=\textwidth]{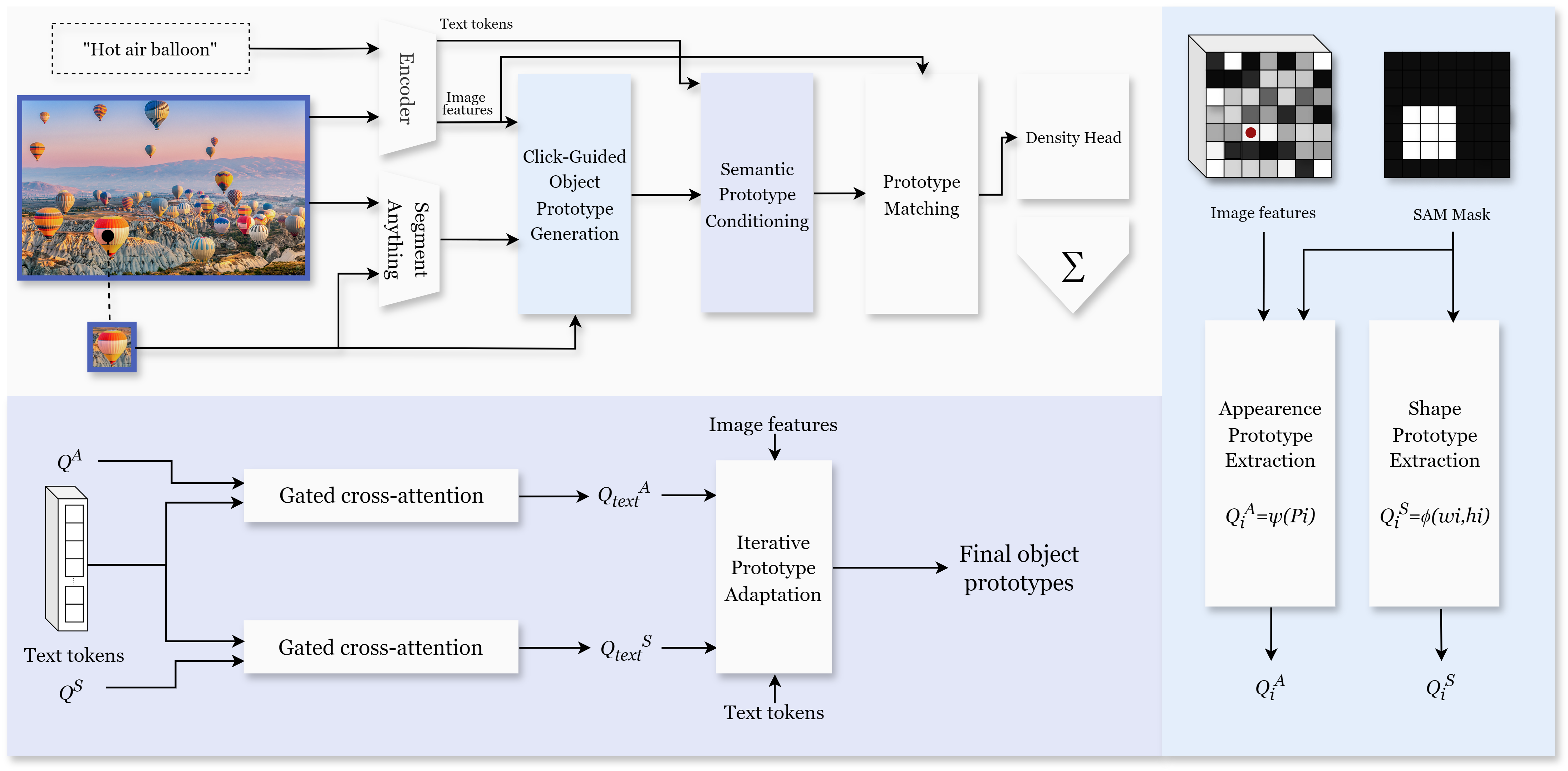}
    \caption{
    Overview of the proposed click-guided multimodal object counting framework.
    The proposed components are highlighted in color. The click-guided
    exemplar construction module (blue) converts user clicks into object masks
    and extracts shape and appearance representations, while the semantic
    conditioning module (purple) injects category-level information into the
    prototypes. The refined multimodal prototypes are matched with image
    features to generate the final density map and object count.
    }
    \label{fig:architecture}
\end{figure*}

Figure~\ref{fig:architecture} presents an overview of the proposed
click-guided multimodal counting framework. The framework aims to construct
strong object prototypes from minimal user interaction by combining
click-guided visual extraction with semantic guidance.

The proposed framework consists of four main stages:
\begin{enumerate}
    \item Visual feature extraction, where the input image is encoded
    into a feature representation used for prototype construction and object
    localization.
    \item Click-guided exemplar construction, where a user-provided
    click is processed by a promptable visual segmentation model to recover the
    complete object region. The resulting mask is used to extract complementary
    shape and appearance prototypes.
    \item Multimodal prototype refinement, where visual prototypes
    are enhanced using semantic information extracted from text descriptions.
    The resulting prototypes are iteratively refined by integrating appearance,
    image context, and semantic cues.
    \item Prototype matching and density prediction, where the refined
    object prototypes are matched with image features to generate the final
    density map and object count.
\end{enumerate}

\subsection{Visual Feature Extraction}

The input image is processed by a pretrained convolutional backbone to extract
multi-scale features, which are resized to a common resolution, concatenated,
and projected into a unified embedding space. A Transformer encoder with
positional encoding then captures long-range interactions between image
regions, producing the encoded visual representation
$F_E\in\mathbb{R}^{d\times h\times w}$ (with embedding dimension $d$ and
spatial size $h\times w$), which serves as visual memory for the subsequent
prototype extraction and refinement stages.

\subsection{Click-guided Exemplar Representation}

The proposed framework introduces a click-based exemplar representation that
transforms sparse user interactions into object-aware visual prompts.

\subsubsection{Object Mask Extraction}

For each exemplar click $c_i$, the framework uses a prompt-based visual
segmentation module to identify the complete object region associated with the
clicked location.

Given the input image and a user interaction point, the segmentation module
predicts an object mask:
\begin{equation}
M_i=\Phi(I,c_i),
\end{equation}
where $M_i$ represents the predicted object region corresponding to the
$i$-th exemplar.

The segmentation module is kept frozen during training and is only used to
extract object masks from sparse user interactions. These masks provide
object-level spatial information that is subsequently used to construct the
exemplar representations.

\subsubsection{Mask-guided Prototype Construction}

The extracted object masks provide object-level spatial information that is used
to construct exemplar representations. For each mask $M_i$, we extract two
complementary representations: a shape prototype describing the object
geometry and an appearance prototype capturing its visual characteristics.

\paragraph{Shape Prototype.}
The foreground region of $M_i$ yields a tight bounding region
$\hat b_i=(x_1,y_1,x_2,y_2)$, given by the minimum and maximum foreground
locations, from which the object width and height are computed as
$w_i=x_2-x_1$ and $h_i=y_2-y_1$. These geometric descriptors are transformed
into a shape prototype $Q_i^S=\phi(w_i,h_i)\in\mathbb{R}^{s^2\times d}$.

\paragraph{Appearance Prototype.}
The object mask is resized to the spatial resolution of the encoded image
features. The mask is then applied to the feature map to retain object-specific
information:
\begin{equation}
F_i^{m}=F_E\odot\tilde M_i .
\end{equation}
The masked object region is extracted using the corresponding bounding region:
\begin{equation}
P_i=\mathrm{RoIAlign}(F_i^{m},\hat b_i),
\end{equation}
while the mask coverage and the original unmasked features are obtained as:
\begin{equation}
C_i=\mathrm{RoIAlign}(\tilde M_i,\hat b_i),
\quad
B_i=\mathrm{RoIAlign}(F_E,\hat b_i).
\end{equation}

The appearance prototype is constructed from the masked feature representation.
Since masking may reduce feature responses in partially covered spatial
locations, we normalize the masked features using the corresponding coverage:
\begin{equation}
Q_i^A(u,v)
=
\frac{P_i(u,v)}
{\max(C_i(u,v),\tau)},
\end{equation}
where $C_i$ represents the foreground coverage ratio and $\tau$ is a small
constant preventing numerical instability.

For locations with insufficient coverage ($C_i(u,v)\leq\tau$), the masked
representation is replaced by the original region feature, $Q_i^A(u,v) =
B_i(u,v)$, reducing the impact of inaccurate mask boundaries while preserving
the benefits of object-focused feature extraction. Since both masked and
unmasked representations come directly from $F_E$, no additional projection
is required, yielding $Q_i^A\in\mathbb{R}^{s^2\times d}$.

\subsection{Semantic Prototype Conditioning}

Although visual prototypes provide instance-level information, appearance cues
alone may be insufficient to capture the semantic identity of the target
category. Therefore, we introduce semantic guidance using a frozen text
encoder.

Given a text description $T$, the text encoder extracts a semantic
representation:
\begin{equation}
z_T = E_{\mathrm{text}}(T),
\end{equation}
which is projected into the prototype embedding space:
\begin{equation}
\tilde z_T = \mathrm{Proj}(z_T),
\end{equation}
where $\mathrm{Proj}(\cdot)$ is a learnable linear projection followed by
Layer Normalization.

To incorporate semantic information while preserving the original prototype
structure, we employ a gated cross-attention mechanism. For an input prototype
$Q$, the semantic conditioning operation is defined as:
\begin{equation}
\mathcal{T}(Q,\tilde z_T)
=
Q+
\tanh(g)\,
\mathrm{MHA}
(
\mathrm{LN}(Q),
\tilde z_T,
\tilde z_T
),
\end{equation}
where $\mathrm{MHA}(\cdot)$ denotes multi-head cross-attention and $g$ is a
learnable scalar gate initialized to zero. This initialization allows the model
to preserve the original visual prototype representation before progressively
learning semantic corrections.

The semantic conditioning is applied independently to the shape and appearance
prototypes: $Q^S_{\mathrm{text}}=\mathcal{T}(Q^S,\tilde z_T)$ and
$Q^A_{\mathrm{text}}=\mathcal{T}(Q^A,\tilde z_T)$.

The resulting semantically enhanced prototypes are then used in the iterative
prototype refinement process. To maintain semantic consistency during visual
reasoning, an additional gated text cross-attention operation is introduced
after appearance and image-context aggregation:
\begin{equation}
Q^{(3)}_{\ell}
=
Q^{(2)}_{\ell}
+
\tanh(g_{\ell})
\,
\mathrm{MHA}
(
\mathrm{LN}(Q^{(2)}_{\ell}),
\tilde z_T^{(\ell)},
\tilde z_T^{(\ell)}
).
\end{equation}
Each refinement layer has an independent learnable gate $g_{\ell}$ and text
projection $\tilde z_T^{(\ell)}$. When semantic information is unavailable,
the residual text branch becomes inactive, allowing the refinement process to
operate using only visual information.

\subsection{Prototype Matching and Density Prediction}

After iterative refinement, the final prototypes $Q_L$ are used to identify
target objects in the image feature space. Each refined prototype is matched
with the encoded image features through a depth-wise correlation operation,
producing exemplar-specific response maps.

The resulting response maps are aggregated and processed by a density
regression decoder to generate the final density map $R$, from which the
predicted count $\hat{N}$ is obtained as in Eq.~(\ref{eq:count}).

The prototype matching and density regression stages remain unchanged in the
overall framework, allowing the proposed improvements to focus on the
click-guided exemplar construction and semantic prototype conditioning
modules.

\subsection{Optimization Objective}

The model is trained using a density regression objective. Given a ground-truth
density map $R^*$, the primary training objective minimizes the difference
between the predicted and target density distributions:
\begin{equation}
\mathcal{L}_{\mathrm{density}}
=
\frac{1}{N_o}
\left\|
R-R^*
\right\|_2^2 ,
\end{equation}
where $R$ and $R^*$ denote the predicted and ground-truth density maps,
respectively, and $N_o$ represents the number of annotated objects.

To improve the intermediate prototype refinement process, auxiliary supervision
is applied to intermediate density predictions. When semantic descriptions are
available, an additional alignment objective encourages the refined prototypes
to preserve consistency with the extracted text representation.

The complete training objective is defined as:
\begin{equation}
\mathcal{L}
=
\mathcal{L}_{\mathrm{density}}
+
\lambda_{aux}\mathcal{L}_{aux}
+
\lambda_{text}\mathcal{L}_{text}.
\end{equation}
The semantic alignment term is activated only when valid text information is
provided. Otherwise, the model is optimized using only the density estimation
objectives.

\section{Experiments}

We evaluate our click-guided counting framework against class-agnostic counting baselines, measure the effect of replacing box exemplars with click-guided representations, and ablate visual exemplar construction and semantic prototype conditioning.
\subsection{Datasets}

\textbf{FSC-147}~\cite{Ranjan2021LearningToCount}.
We use the FSC-147 dataset as the primary benchmark for class-agnostic object
counting. The dataset contains 6,135 images covering 147 object categories.
Following the standard protocol, the categories are divided into
89 training categories, 29 validation categories, and 29 test categories.
Each image is originally annotated with three exemplar bounding boxes that
define the reference objects for counting.

In our framework, each exemplar box is replaced by a single click, used as a
SAM prompt to generate an object mask.

\begin{table*}[t]
\centering
\caption{Comparison with state-of-the-art methods on FSC-147 object counting under different prompt settings.}
\label{tab:fsc147_comparison}
\small
\setlength{\tabcolsep}{4pt}
\renewcommand{\arraystretch}{1.1}
\begin{tabular}{llcccc}
\toprule
\textbf{Prompt} & \textbf{Method} & \textbf{Validation MAE} & \textbf{Validation RMSE} & \textbf{Test MAE} & \textbf{Test RMSE} \\
\midrule
\multirow{7}{*}{box}
 & FR \cite{Kang2019FeatureReweighting}       & 45.45 & 112.53 & 41.64 & 141.04 \\
 & FSOD \cite{Fan2020AttentionRPN}            & 36.36 & 115.00 & 32.53 & 140.65 \\
 & C-DETR \cite{Nguyen2022CountingDETR}       & --    & --     & 16.79 & 123.56 \\
 & SAM \cite{Kirillov2023SAM,Ma2023SAMCounting} & 31.20 & 100.83 & 27.97 & 131.24 \\
 & SAM-Free \cite{Shi2024TFPOC}               & --    & --     & 19.95 & 132.16 \\
 & LOCA \cite{Dukic2023LOCA}                  & 10.24 & 32.56  & 10.79 & 56.97 \\
 & PSeCo \cite{Huang2024PseCo}                & 15.31 & 68.34  & 13.05 & 112.86 \\
\midrule
\multirow{5}{*}{text}
 & ZSOC \cite{Xu2023ZSOC}                     & 26.93 & 88.63 & 22.09 & 115.17 \\
 & CLIP-C \cite{Jiang2023CLIPCount}           & 18.79 & 61.18 & 17.78 & 106.62 \\
 & TFPOC \cite{Shi2024TFPOC}                  & --    & --    & 24.79 & 137.15 \\
 & UPBC \cite{Lin2024UPBC}                    & 16.92 & 58.92 & 16.81 & 105.83 \\
 & GroundingDINO \cite{Liu2023GroundingDINO,Liu2025CountSE}
 & 12.07 & 57.02 & 11.52 & 97.37 \\
\midrule
\multirow{3}{*}{point}
 & TFPOC \cite{Shi2024TFPOC}      & --    & --    & 20.10 & 132.83 \\
 & UPBC \cite{Lin2024UPBC}        & 17.16 & 59.38 & 15.86 & 103.27 \\
 & \textbf{Ours}
 & \textbf{11.57} & \textbf{35.69}
 & \textbf{12.31} & \textbf{89.40} \\
\midrule
\textbf{point + text} & \textbf{Ours}
 & \textbf{9.95} & \textbf{36.03}
 & \textbf{9.77} & \textbf{68.69} \\
\bottomrule
\end{tabular}
\end{table*}

\noindent\textbf{COCO-derived counting benchmark.}
To evaluate the generalization capability of class-agnostic counting models under
category shifts, we construct an additional counting benchmark from the COCO 2017
instance annotations. Unlike the standard COCO evaluation protocol, where the
same categories may appear across training and evaluation images, we adopt a
category-disjoint setting in which evaluation categories are completely
excluded from training.

Specifically, 20\% of the available COCO categories are randomly selected as
held-out categories and removed from the training split. The remaining
categories are used for training, while validation and test samples are created
only from the held-out category pool. The held-out samples are further divided
according to image identities to ensure that evaluation is performed on unseen
categories.

To construct counting samples, we consider only non-crowd object instances and
retain images containing at least three objects from the same category. Object
instances are filtered using minimum area and bounding-box size constraints to
remove extremely small or unreliable annotations. For each image-category pair,
three object instances are selected as exemplars, and their spatial locations
are converted into point-based guidance for the proposed framework. The
resulting benchmark evaluates whether a model can adapt its learned object
representation to categories that are not observed during training.

The dataset statistics are summarized in Table~\ref{tab:datasets}.

\begin{table}[t]
\centering
\caption{Statistics of the datasets used for evaluation.}
\label{tab:datasets}
\small
\setlength{\tabcolsep}{4pt}
\renewcommand{\arraystretch}{1.15}
\resizebox{\columnwidth}{!}{
\begin{tabular}{lcc}
\toprule
\textbf{Dataset Property} & \textbf{FSC-147} & \textbf{COCO-derived} \\
\midrule
Images  & 6,135  & 66,823 \\
Categories  & 147  & 80 \\
Split protocol  & Category-disjoint  & Category-disjoint \\
Instances per sample  & $\geq$ 3  & $\geq$ 3 \\
Evaluation task  & Class-agnostic counting  & Class-agnostic counting \\
\bottomrule
\end{tabular}
}
\end{table}

\subsection{Experimental Setup}

The visual encoder uses a ResNet-50 backbone pretrained with SwAV; multi-scale
features are projected into a common embedding space and processed by a
3-layer Transformer encoder with positional encoding to produce the image
memory. Images are resized to $512\times512$ with a feature reduction factor
$\rho=8$; the prototype refinement module uses embedding dimension $d=256$,
8 attention heads, kernel size $s=3$, and three refinement steps. Object masks
are precomputed offline from click annotations to accelerate training and
inference.

The model is optimized with AdamW ($\text{lr}=10^{-4}$, weight decay
$10^{-4}$) for 200 epochs with batch size 4 per GPU on two NVIDIA GPUs, with
the learning rate reduced after epoch 120 and gradient clipping at norm 0.1.
The auxiliary loss weight is 0.3; when semantic guidance is used, the text
alignment loss weight is 0.1, activated after a 30-epoch warm-up.

\subsection{Evaluation Metrics}

Following previous class-agnostic counting works, we evaluate the model using
Mean Absolute Error (MAE), Root Mean Squared Error (RMSE), and Normalized
Absolute Error (NAE):
\begin{equation}
MAE=
\frac{1}{|D|}
\sum_{i\in D}
|\hat{N_i}-N_i|,
\end{equation}
\begin{equation}
RMSE=
\sqrt{
\frac{1}{|D|}
\sum_{i\in D}
(\hat{N_i}-N_i)^2
},
\end{equation}
\begin{equation}
NAE=
\frac{1}{|D|}
\sum_{i\in D}
\frac{|\hat{N_i}-N_i|}{N_i},
\end{equation}
where $\hat{N_i}$ represents the predicted object count obtained by integrating
the predicted density map, and $N_i$ denotes the ground-truth object count.
The dataset size is represented by $|D|$.

\subsection{Experimental Results}

\textbf{Results on FSC-147.}
We evaluate our framework on the FSC-147 benchmark, which provides a
challenging setting for class-agnostic object counting with diverse object
categories. Table~\ref{tab:fsc147_comparison} reports comparisons with existing
methods under different prompt modalities, including box, text, and point-based
guidance.

The results demonstrate that click-based object representations offer an
effective alternative to manually annotated bounding boxes, achieving
competitive performance with box-supervised approaches while substantially
reducing annotation burden. Incorporating semantic information through text
conditioning provides further improvements over point guidance alone: the
combination of point and text prompts achieves the lowest MAE among all compared methods, showing that instance-level visual cues and category-level
semantics are complementary for prototype refinement. However, this improvement is uneven across metrics: while point+text achieves the lowest MAE overall, its RMSE remains higher than the box-supervised LOCA baseline (68.69 vs. 56.97 on the test set), suggesting that click-derived prototypes are more prone to large errors on a subset of challenging samples even as they improve average-case accuracy.

\noindent\textbf{Results on the COCO-derived benchmark.}
On the category-disjoint COCO-derived benchmark, our model achieves MAE of
1.57/1.48 and RMSE of 2.23/2.11 on validation/test (Table~\ref{tab:coco_results}).
The close agreement between splits indicates stable prototype construction
under category shifts, transferring counting ability to categories unseen
during training.

Because COCO-derived images contain fewer instances per image than FSC-147,
raw MAE/RMSE are not directly comparable across datasets; we instead compare
normalized absolute error (NAE). Similar NAE values across the two benchmarks
(0.23/0.26 vs.\ 0.27/0.28 for validation/test) indicate consistent relative
counting accuracy despite different count distributions.

\begin{table}[t]
\centering
\caption{Performance of the click-based (point-only) C\&C model on FSC-147
and the proposed category-disjoint COCO-derived counting benchmark, evaluated
without semantic text guidance.}
\label{tab:coco_results}
\small
\setlength{\tabcolsep}{4pt}
\renewcommand{\arraystretch}{1.15}
\begin{tabular}{lccc}
\toprule
\textbf{Dataset} & \textbf{MAE} & \textbf{RMSE} & \textbf{NAE} \\
\midrule
FSC-147 (point)
& 11.57 / 12.31
& 35.69 / 89.40
& 0.23 / 0.26 \\
COCO-derived (point)
& 1.57 / 1.48
& 2.23 / 2.11
& 0.27 / 0.28 \\
\bottomrule
\end{tabular}
\end{table}

\subsection{Ablation Study}

Table~\ref{tab:ablation} isolates the two proposed components on FSC-147.
Replacing box exemplars with one click per exemplar (Ours) costs +1.33/+1.52
MAE (val/test) and raises test RMSE from 56.97 to 89.40. Adding text
(Ours + Text) lowers MAE (11.57 to 9.95 val, 12.31 to 9.77 test) and test RMSE
(89.40 to 68.69), bringing MAE below LOCA's; however, validation RMSE slightly
increases (35.69 to 36.03) and test RMSE stays above LOCA's.

\begin{table}[t]
\centering
\caption{Ablation of exemplar type and text guidance: box exemplars (LOCA) vs.\ click-only vs.\ click+text (ours).}
\label{tab:ablation}
\small
\setlength{\tabcolsep}{4pt}
\renewcommand{\arraystretch}{1.1}
\begin{tabular}{lcccc}
\toprule
\textbf{Method} & \textbf{Val MAE} & \textbf{Val RMSE} & \textbf{Test MAE} & \textbf{Test RMSE} \\
\midrule
LOCA (box) & 10.24 & 32.56 & 10.79 & 56.97 \\
Ours (point) & 11.57 & 35.69 & 12.31 & 89.40 \\
\textbf{Ours + Text} & \textbf{9.95} & \textbf{36.03} & \textbf{9.77} & \textbf{68.69} \\
\bottomrule
\end{tabular}
\end{table}

\section{Conclusion}

In this paper, we present C\&C, a click-guided class-agnostic counting framework that converts sparse clicks into object-aware exemplars, eliminating bounding-box annotations. Semantic prototype conditioning further integrates textual information during refinement. Experiments on FSC-147 and a category-disjoint COCO-derived benchmark demonstrate its effectiveness and generalization to unseen categories. Future work will explore richer vision-language representations
and more adaptive multimodal prototype learning.

\section*{Acknowledgment}

This work was supported by the Canada Research Chair in Artificial Intelligence for Suicide Prevention (CRC-2023-00036).

\bibliographystyle{IEEEtran}
\bibliography{references}

\end{document}